\documentclass[sigconf]{acmart}

\usepackage[utf8]{inputenc} 
\usepackage[T1]{fontenc}    
\usepackage{hyperref}       
\usepackage{url}            
\usepackage{booktabs}       

\usepackage{algpseudocode}
\usepackage{algorithm}
\usepackage{graphicx}
\usepackage{subfig}
\usepackage{textcomp}
\usepackage{gensymb}
\usepackage{epsfig} 
\usepackage{xcolor}

\usepackage{booktabs}
\usepackage{array}
\usepackage{multirow}
\usepackage{caption}

\usepackage{balance}

\setcopyright{acmcopyright}
\copyrightyear{2022}
\acmYear{2022}
\acmDOI{10.1145/nnnnnnn.nnnnnnn}

\acmConference[GECCO '22]{The Genetic and Evolutionary Computation Conference 2022}{July 9--13, 2022}{Boston, USA}
\acmISBN{978-x-xxxx-xxxx-x/YY/MM} 

\begin{document}

\title{Assessing Evolutionary Terrain Generation Methods for Curriculum Reinforcement Learning}

\author{David Howard}
\orcid{0002-5012-7224}
\affiliation{
  \institution{CSIRO}
  \city{Brisbane}
  \state{Queensland}
  \country{Australia}}
\email{david.howard@csiro.au}

\author{Josh Kannemeyer}
\affiliation{
  \institution{Monash University}
  \streetaddress{Clayton}
  \city{Melbourne} 
  \state{Victoria} 
  \country{Australia}}

\author{Davide Dolcetti}
\affiliation{
  \institution{Queensland University of Technology}
  \streetaddress{Brisbane City}
  \city{Brisbane} 
  \state{Queensland} 
  \country{Australia}}
  
\author{Humphrey Munn}
\affiliation{
\institution{University of Queensland}
  \streetaddress{St Lucia}
  \city{Brisbane} 
  \state{Queensland} 
  \country{Australia}}

\author{Nicole Robinson}
\affiliation{
  \institution{Monash University}
  \streetaddress{Clayton}
  \city{Melbourne} 
  \state{Victoria} 
  \country{Australia}}

\begin{abstract}
Curriculum learning allows complex tasks to be mastered via incremental progression over `stepping stone' goals towards a final desired behaviour. Typical implementations learn locomotion policies for challenging environments through gradual complexification of a terrain mesh generated through a parameterised noise function. To date, researchers have predominantly generated terrains from a limited range of noise functions, and the effect of the generator on the learning process is underrepresented in the literature. We compare popular noise-based terrain generators to two indirect encodings, CPPN and GAN. To allow direct comparison between both direct and indirect representations,  we assess the impact of a range of representation-agnostic MAP-Elites feature descriptors that compute metrics directly from the generated terrain meshes.  Next, performance and coverage are assessed when training a humanoid robot in a physics simulator using the PPO algorithm.  Results describe key differences between the generators that inform their use in curriculum learning, and present a range of useful feature descriptors for uptake by the community.
\end{abstract}



\begin{teaserfigure}
\centering
  \includegraphics[width=0.8\textwidth]{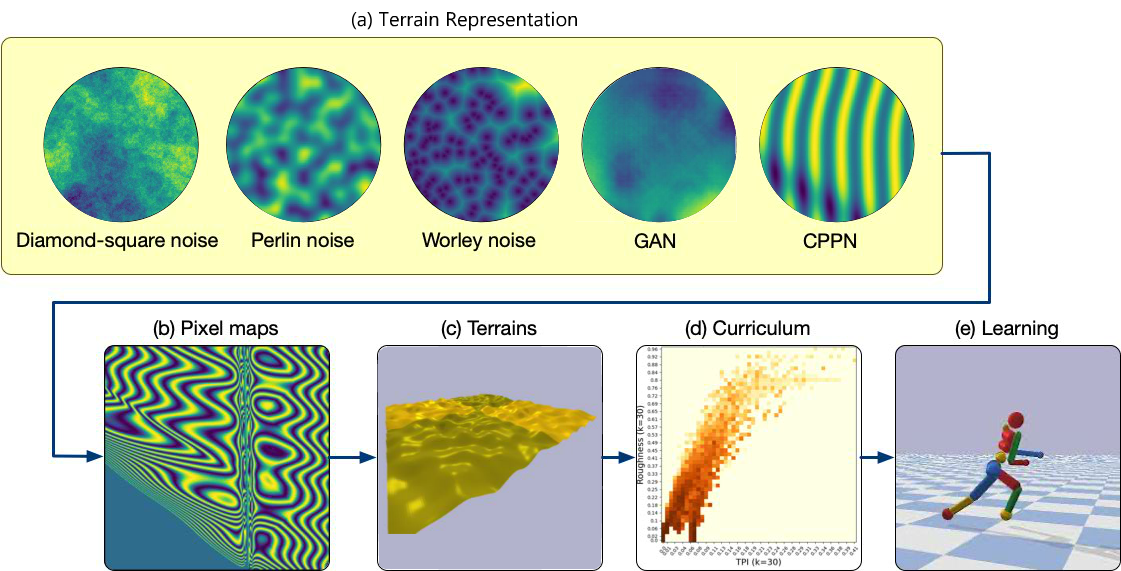}
  \caption{System overview: (a) Evolved terrain representations populate (b) 2D pixel maps and are subsequently turned into (c) terrain meshes which fill a (d) MAP-Elites archive.  A bipedal walker is subsequently trained via PPO (e) and its learning performance is used to assess the impact of the various generators.}
  \label{fig:fig1}
\end{teaserfigure}
\maketitle

\section{Introduction}

Curriculum Learning (CL) \cite{narvekar2020curriculum,bengio2009curriculum, TiddBrendan2020GCLf} is a powerful reinforcement learning technique that engenders powerful behaviours by gradually increasing the difficulty of the task to be solved.  This incremental approach has numerous benefits including performance and generalisation ability. To maximise the benefit of a curriculum, it is important to have a training environment that increases in complexity proportional to the learning ability of the agent, and contains relevant features to encourage learning.  Diverse training examples, presented to the learner in an appropriate order \cite{heess2017emergence}, have resulted in the generation of capable, complex behaviours for simulated robots \cite{TiddBrendan2020GCLf, xie2020allsteps}, and aided sim2real transfer of those policies to real robot deployments \cite{lee_learning_2020}.

A typical setup for learning locomotion skills (a popular focus area) involves the creation of a set of terrains of varying difficulty through some terrain generator, typically a parameterised noise function.  The terrains are then presented to the agent, generally in order of difficulty, and the agent learns by solving simpler examples to eventually reach a given target behaviour.  Noise functions are popular candidate terrain generators, as increasing terrain difficulty can be simply realised by increasing values of the noise parameters.  Different generators produce terrains with different geometric features, which has an affect on learning.  

To date, the impact of the chosen terrain generator has not been explored, and this is the focus of our paper.  We address the effects of terrain generator on the learning process in two steps.  In step 1, we select a varied range of popular terrain generators from the literature, which includes both direct and indirect encodings. Because the difficulty of an indirectly-coded terrain cannot be ascertained {\em a priori}, we categorise the resulting terrain mesh according to a set of representation-agnostic features selected from related literature.  The selected features become feature descriptors in MAP-Elites, providing a diverse set of high-quality terrains.  Terrains are evolved to fill the archive and generate one curriculum per generator type.   In step two, each curriculum is used to train a bipedal walker, and results compare reachable terrain difficulty and learning speed.  Our approach is illustrated in Figure \ref{fig:fig1}.

We present two main original contributions; (i) the selection of appropriate representation-agnostic feature descriptors, including coverage analysis of curricula evolved using those features, and (ii) analysis of the effects of generator type on learning performance and rate.  Results show the identification of suitable feature descriptors for representation-agnostic terrain-based curriculum learning, and suggest key differences between the  generators, particularly between direct and indirect representations in terms of map coverage.  We provide evidence that a multi-generator approach may be beneficial to the generation of curricula that promote rapid learning.  

The remainder of the paper is organised as follows; Section 2 presents pertinent background research.  Section 3 describes our methodology.  Section 4 presents experiments and results, and Section 5 provides a discussion.

\section{Background}

Reinforcement Learning (RL) \cite{kober2013reinforcement} is one of the most commonly used methods to train agent behaviours, and is similar conceptually to an evolutionary algorithm which learns by continually interacting with (and being rewarded by) an environment.  In this study we use Proximal Policy Optimisation (PPO) \cite{ppo} due to its ubiquity as a benchmark method to train agents in complex environments. PPO is a policy gradient method that alternates between sampling data through the agent's interaction with the environment and optimising an objective function using stochastic gradient descent. PPOs main advantage over other policy gradient techniques is the use of multiple epochs of mini-batch updates rather than performing one gradient update per data sample.

Curriculum learning \cite{bengio2009curriculum} is a form of incremental learning \cite{howard2014evolving}, designed to address some common challenges with RL \cite{narvekar2020curriculum}.  Firstly, it allows for agents to solve very hard problems by progressively building competency in easier versions of the problem  \cite{bongard_utility_2010}.  Secondly, it provides a wealth of learning experiences to the agent which can greatly assist with generalisation \cite{KadianAbhishek2020SPDE, HoferSebastian2021SiRa,nichol2018gotta}.   Lee et al. \cite{lee_learning_2020} demonstrate both of these benefits using a particle filter to update parameters of a noise function-based terrain generator, allowing complex terrains to be navigated by a real quadruped after simulated learning.

A guided curriculum approach \cite{TiddBrendan2020GCLf} incrementally removed supporting forces from a bipedal walker's body before increasing terrain complexity and diversity.  Accelerated learning of complex environments is also evidenced with hexapod robots~\cite{9043822}. Miras demonstrated that balancing behaviour could be achieved by training on a tilted plain rather than a flat plain with objects on it, even though balance was not incorporated into the fitness function of the robot~\cite{miras_effects_2019}. Huizinga \cite{huizinga2019evolving} found that ordering of sub-tasks towards learning a difficult task is challenging, and instead proposed the Combinatorial Multi-Objective Evolutionary Algorithm (CMOEA) to simultaneously explore all orderings. This is an effective CL method when subtasks can be clearly defined (e.g. jumping, walking). Xie \cite{xie2020allsteps} demonstrated the critical role of a curriculum to train three bipedal agents to walk on stepping-stone scenarios, with final terrain complexity and learning rate being superior for curriculum approaches compared to non-curriculum RL.  Akkaya \cite{akkaya2019solving} presents a CL approach using automatic domain randomization (ADR), demonstrating vastly improved sim2real transfer compared to a non-curriculum baseline. ADR automatically expands the randomisation range parameterising a distribution over environments. Florensa \cite{florensa2018automatic} showed that using a curriculum generating network it is possible to train an agent to `perform a wide set of tasks without requiring any prior knowledge of its environment'.   Open-ended curricula \cite{zhou2021openended} can learn on terrains that are continually generated during agent learning \cite{poet}.

Procedural Content Generation (PCG) originated in computer graphics and video game design. PCG can create large volumes of high-quality content with controllable randomness, including digital objects, landscapes, levels, textures and 3D models.  PCG has been readily adopted by the machine learning community ~\cite{risi2020increasing} to create a wealth of training data.  Rich and diverse environments encourage agent learning and improve robustness \cite{heess2017emergence}, with works showing that curriculum learning can train several quadrupedal robot policies in parallel to walk over uneven terrains in simulation \cite{rudin2021learning}. Terrain generators have been previously evolved \cite{evoTG,asasas}, however these works focus only on a single representation and do not evaluate terrains in an agent-based learning context.  Also in an evolutionary context, quality-diversity algorithms \cite{pugh2016quality} present an effective method for storing and traversing a curriculum, shown in both robotics and game-playing contexts \cite{Cully_2015} \cite{9300206}.  In particular,the use of aligned feature dimensions within MAP-Elites \cite{mouret2015illuminating} can provide direction to a (stochastic) traversal algorithm. 

Overall, we see that curriculum learning is a powerful technique relevant to both evolutionary algorithms and reinforcement learning.  The literature abounds with a smorgasbord of both direct and indirectly-represented generators, including various noise functions as well as CPPNs and GANs.  Additionally, we see many different feature dimensions being used to store the curriculum and permit traversal. However, we identify a significant literature gap in that these different selections of generators and features have not yet been compared.  In this study we compare popular generators and a set of selected feature descriptors for an evolved MAP-Elites based curriculum.  We attempt to inform the selection of appropriate terrain generators and feature dimensions for fellow researchers in the field.  We also demonstrate a curriculum learning approach that simultaneously supports both direct and indirect representations, opening up future work in mixed-generator curricula.

\section{Methodology}

We follow a three-stage process: (i) decide on features, (ii) evolve curricula (iii) learn on curricula.  To allow for the isolated study of generators, we omit other tasks that typically comprise a curriculum (steps, jumps, etc.) and focus purely on locomotion in rough terrains.

\subsection{Generators}

We select 5 popular and diverse generators from the literature: 3 direct (Perlin noise, Diamond Square Noise, Worley Noise) and two indirect (a GAN, and a CPPN). Each generator mapped to a 256x256 resolution pixel map which is converted to a heightmap and then to a solid terrain mesh for simulation. 

\subsubsection{Perlin Noise}

Perlin noise was applied to fractal Brownian Motion (fBM). The genome was parameterised into scale [1-100], octaves [1-9], persistence [0.1-0.9], lacunarity [1-3] and a random seed [0-100].

\subsubsection{Diamond Square Noise}

We use a version of the diamond square noise algorithm \cite{diamondsq}.  The four corner points of the heightmap grid are initialised with a random value between -1 and 1 making a square. The following two steps alternate until all the values of the heightmap are assigned:

Diamond step: for each square in the grid, assign the value of the midpoint of that square to be a random percentage \emph{P} of the average of the four corner points, plus a random value \emph{R} between -1 and 1.

Square step:  for each diamond in the grid, assign the value of the midpoint of the diamond to be a random percentage \emph{P} of the average of the four corner points, plus a random value \emph{R} between -1 and 1.

After each iteration of these steps, the range of the random value \emph{R} is reduced using the formula:

\[
\frac{-1}{(steps*D+1)} \leq R \leq \frac{1}{(steps*D+1)}
\]

With $\emph{D} \in [1,10]$, and \emph{steps} corresponding to the number of iterations of the above two steps performed.  \emph{P} is computed as:

\[
Z\leq P \leq 100-Z, Z \in [0,50]
\]

The genome is represented by a random seed, the four initial corner values, percentile (\emph{Z}), and level (\emph{D}).

\subsubsection{Worley Noise}

Worley Noise is typically used to generate procedural textures, and is implemented following \cite{worley}. The genome for the Worley terrain generator consists of a seed for randomisation, the number of feature points $\emph{N} \in [2,400]$ and the index of the point sorted by ascending distance to the current point $\emph{D} \in [0, \frac{N}{4}]$.

\subsubsection{GAN}

The Deep Convolutional GAN (DCGAN) model was trained on real terrain heightmap data with identical resolution to the humanoid in simulation. The original data was in pointcloud format obtained from the OpenTopography website. The ground plane was semantically segmented, and the resulting points interpolated into multiple heightmap patches at the desired resolution. This provided a large amount of high resolution data (30000 256x256 patches) which was used to train the model. The latent vector fed into the generator contained 50 values which formed the genome of the GAN generator.

\subsubsection{CPPN}

The CPPN terrain generator has 2 inputs corresponding to row and column of the heightmap, and one output for the height value. All remaining CPPN settings were taken from the PicBreeder paper \cite{picbreeder}.

\subsection{Features}

We explore a range of generic terrain descriptors for use as MAP-Elites feature descriptors that allow both direct and indirect generators to be compared. Features were aligned to terrain traversability to provide continuity to the final generated curricula, with enough variation to create diversity in the population.  Features originate from a range of non-RL, non-evolutionary fields, mainly from surveying where they are used to categorise real terrains. We use:

\begin{enumerate}
    \item Terrain Ruggedness Index (TRI) \cite{tri} - measures the average difference between a pixel and its 8 neighbouring pixels.
    \item Topographic Position Index (TPI) \cite{tpi} - measures the difference between each pixel and the mean of its 8 neighbouring pixels
    \item Roughness \cite{roughness} - measures the maximum difference between a pixel and its 8 neighbouring pixels 
    \item Traversability Estimation Model implemented in \cite{traversability} - A trained model that predicts the traversability of terrains by outputting a traversability map. This model had been trained on different types of terrains to the terrain generators used in this paper. The orientation input to the model was set to 0 as this was the direction the humanoid traversed.
\end{enumerate}

Once generated, a terrain mesh is tagged with values for each feature descriptor.  Settings for kernel size are determined in Section 4.  Roughness, TPI and TRI use a stride length of 2.  The average of the output from each feature was taken as the overall measure.  Initial experimentation found that an archive discretisation of 50 bins per descriptor presented meaningful but achievable differences between terrains in neighbouring cells. 

\subsection{Evolutionary Algorithm}

\begin{figure}[h!]
\centering
\includegraphics[width=1\columnwidth]{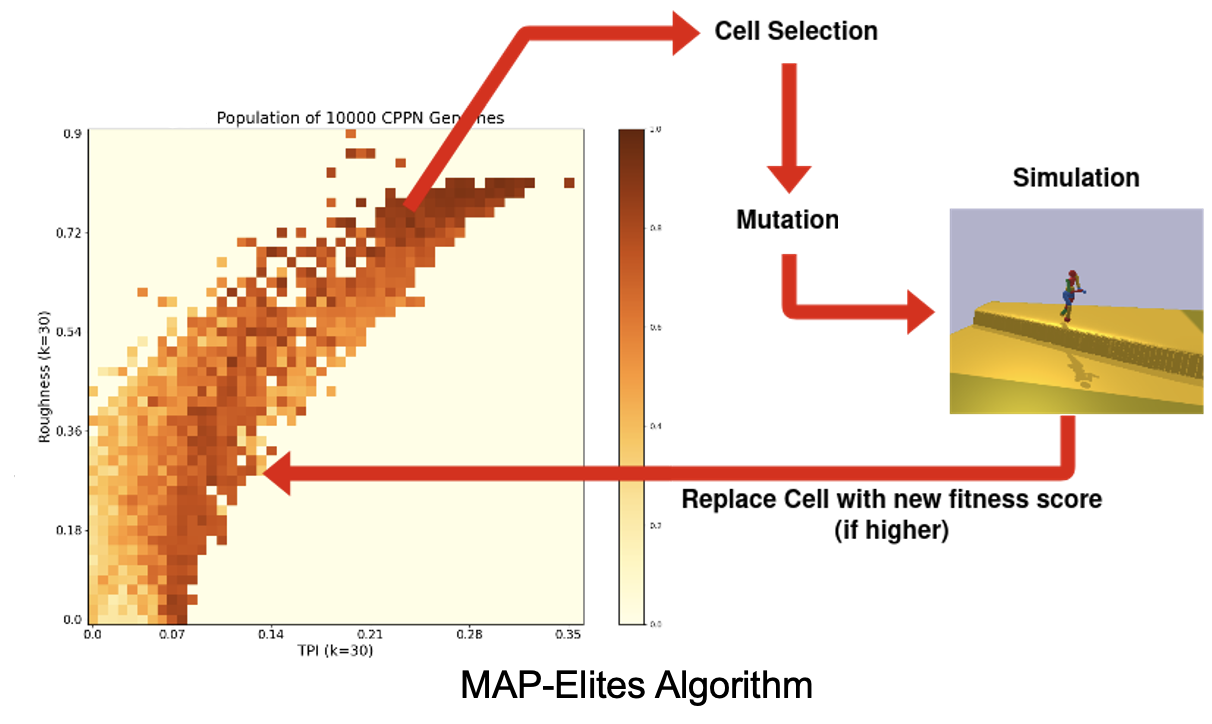}
\caption{Overview of library generation.  Features must be calculate in simulation before the terrain can be added to the library.}
\label{fig:fig2}
\end{figure}

We evolve curriculums for each generator, and compare different pairs of feature dimensions.  Per treatment, we run MAP-Elites for 5000 generations, with 100 random initial genomes.  Per generation, 20 new terrains were generated by randomly selecting from the current members of the archive, mutating, generating feature values, and adding back into the archive as in Figure \ref{fig:fig2}.

For noise emitters, each gene in the genome had a mutation probability of 0.35, and mutation altered the allele by $\pm$ a value taken from a normal distribution with covariance set to 10\% of its range.  CPPN mutation followed NEAT \cite{picbreeder}.  The GAN's latent space was stored in 50 variables, which with P=0.07 were mutated by a value taken from a normal distribution with covariance set to 10\% of its range (heuristically determined).

\begin{figure}[ht!]
\begin{center}
\centering 
 \subfloat[]{\includegraphics[width=0.3\columnwidth]{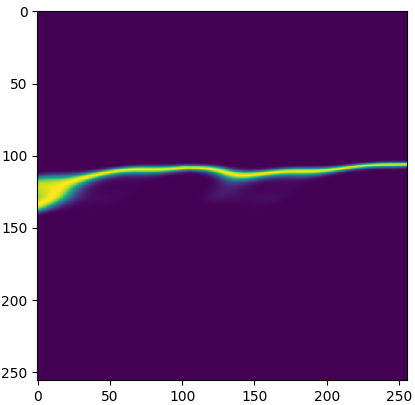} }
 \subfloat[]{\includegraphics[width=0.3\columnwidth]{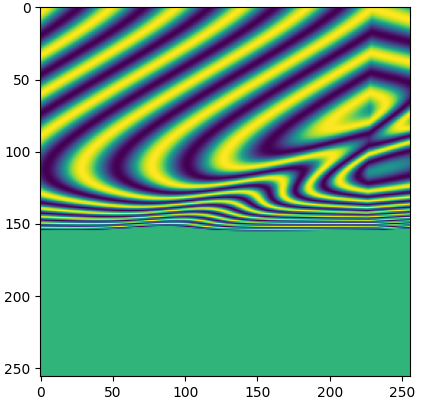} }
 \subfloat[]{\includegraphics[width=0.3\columnwidth]{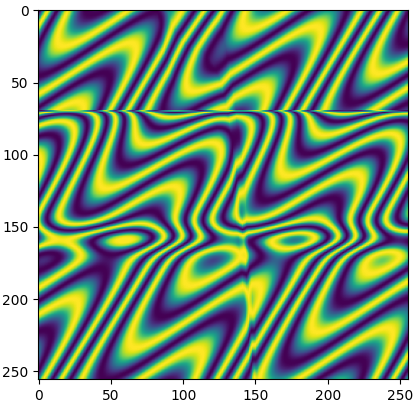} }\\
 \subfloat[]{\includegraphics[width=0.3\columnwidth]{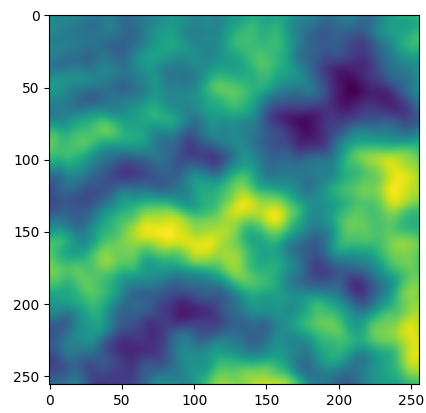} }
 \subfloat[]{\includegraphics[width=0.3\columnwidth]{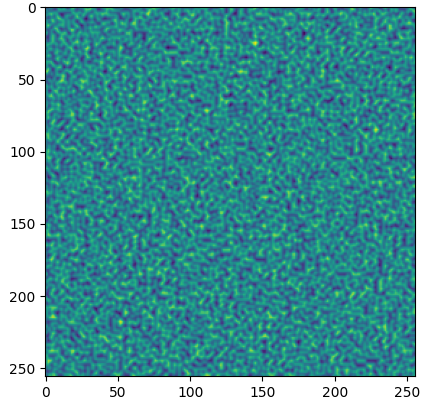} }
 \subfloat[]{\includegraphics[width=0.3\columnwidth]{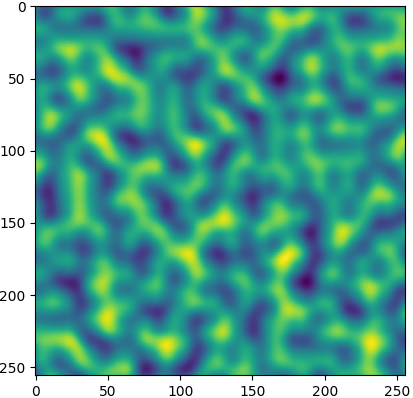} }\\
\end{center}
\caption[]{Representative pixel maps with approximate difficulties of 0.1, 0.5, 0.9 respectively (L-R) for (a)-(c) an indirect representation (CPPN) and (d)-(f) a direct representation (Perlin noise).}
\label{fig:fig3}
\end{figure}

Fitness is set to 1.0 - the difficulty of the generated terrain.  Difficulty is assessed using the base policy of the Bipedal Walker that is trained on flat ground\cite{coumans2017pybullet} and simulated in PyBullet. Difficulty of a terrain is calculated as the average distance travelled out of the best 5 of 20 attempts with random joint initialisations, normalised between 0 and 1. The 5 best were taken as some initialisations were too extreme and caused a large amount of noise in the fitness estimation.  When replacing individuals in a MAP-Elites cell, we kept the lowest difficulty terrain and deleted the higher difficulty one.  This generated a smoother curriculum in terms of fitness, whilst also removing many impossible terrains.  A second step removed the remaining impossible terrains, following Algorithm 1.  Representative terrains for two generators (CPPN and Perlin noise) that achieve difficulties of $\approx$0.1, $\approx$ 0.5, and $\approx$ 0.9 are shown in Figure \ref{fig:fig3}.

\begin{algorithm}[h!]
\caption{Check terrain traversability.}
\begin{algorithmic}
\Procedure{Check Terrain}{$hm$}\Comment{Terrain heightmap}
\State $threshold\gets \frac{robot Height}{3}$\Comment{max incline in metres}
\State $k\gets 26$\Comment{Approx. step length of robot}
\State Initialise  differences result [hm.rows-k,hm.cols-k]
\For{r in [0, hm.rows - kSize)}
\For{c in [0, hm.cols - kSize)}
\State differences[r,c]$\gets$

maximum\_difference(hm[r:r+k,c:c+k])
\EndFor
\EndFor
\State \textbf{return} $max(differences) \leq threshold$
\EndProcedure
\end{algorithmic}
\end{algorithm}

\section{Experimentation}
\subsection{Experiment 1: Features and Curriculum}

\begin{figure}[ht!]
\centering 
\includegraphics[width=0.95\columnwidth]{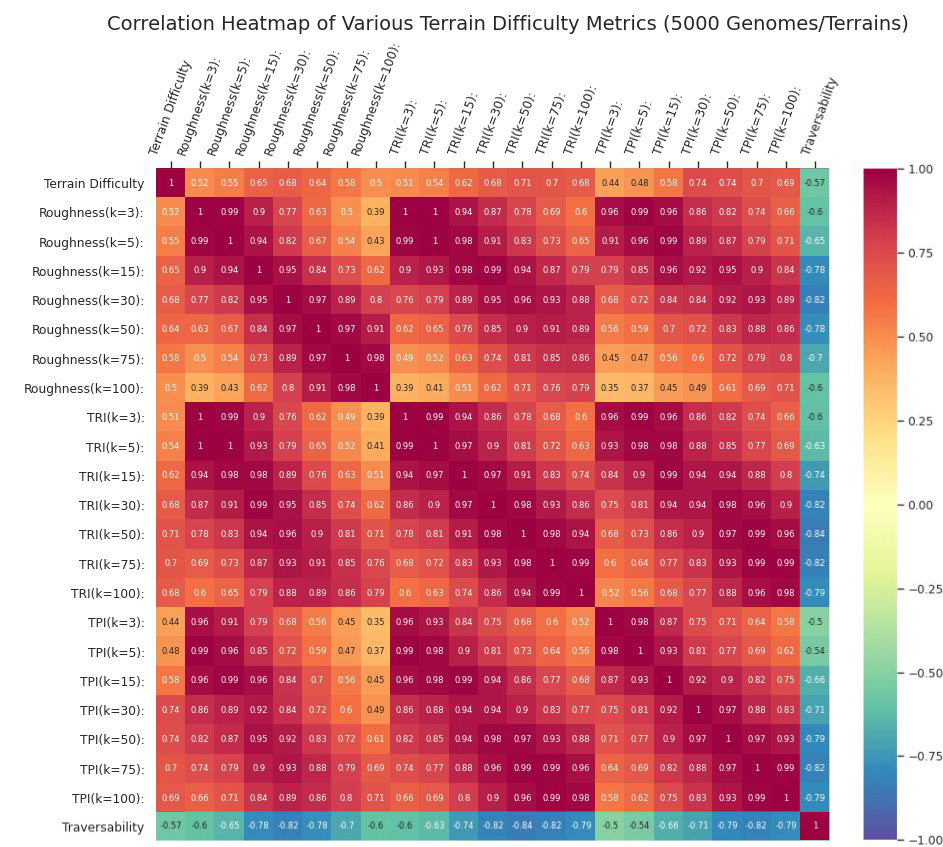}
\includegraphics[width=0.95\columnwidth]{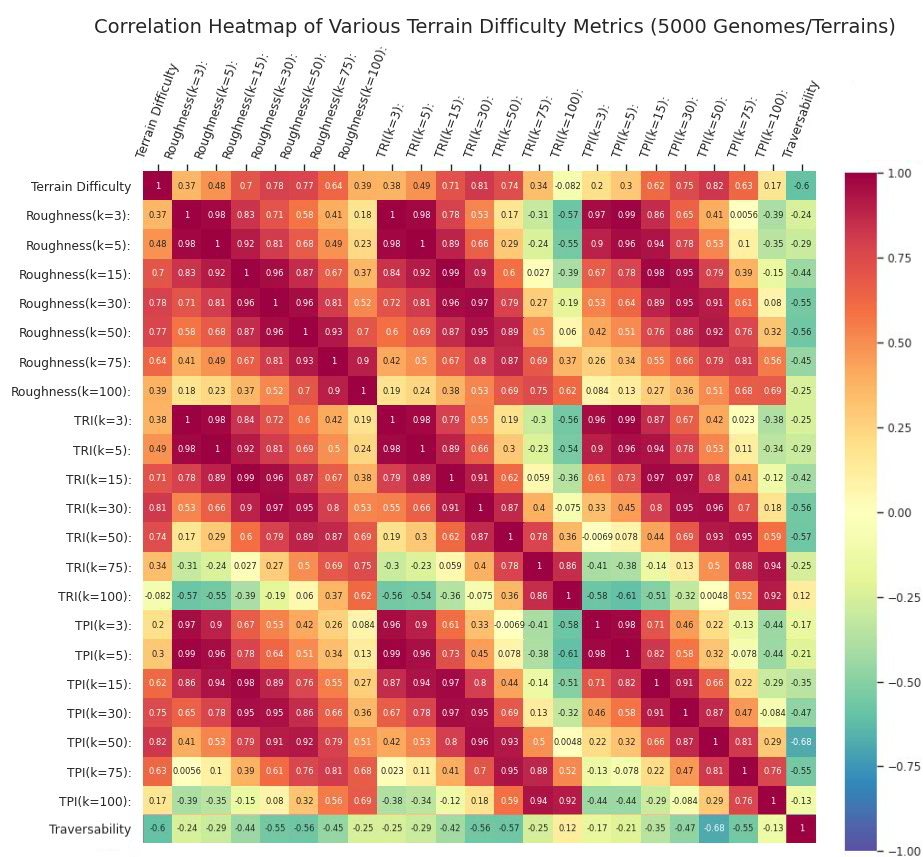}
\caption[]{Heatmaps showing Pearson correlation of all considered terrain characterisation features including ground truth: terrain difficulty, for (a) CPPN and (b) Perlin noise. }
\label{fig:fig4}
\end{figure}

Initial experimentation compared feature descriptors with different kernel sizes to measure their alignment to terrain difficulty (fitness). This alignment was ideal for constraining the solution search space to produce a strong traversability gradient in the MAP-Elites grid. Without a strong gradient, creating an effective curriculum becomes difficult. Roughness, TPI and TRI had similar results with TPI being most correlated. Results are shown in Figure \ref{fig:fig4}. We used TPI in learning experiments as it was most aligned, and use Roughness since TPI and TRI (k = 30) were too similar (r = 0.94/0.95) to produce sufficient map diversity. A kernel size of 30 was chosen as it had consistently high correlations to terrain difficulty.  These features could then be used for arbitrary meshes with confidence that they were meaningful metrics.  The traversability model ("Traversability") had the weakest correlation to difficulty so this was not used further.

\begin{figure}[h]
\centering
\includegraphics[width=0.9\columnwidth]{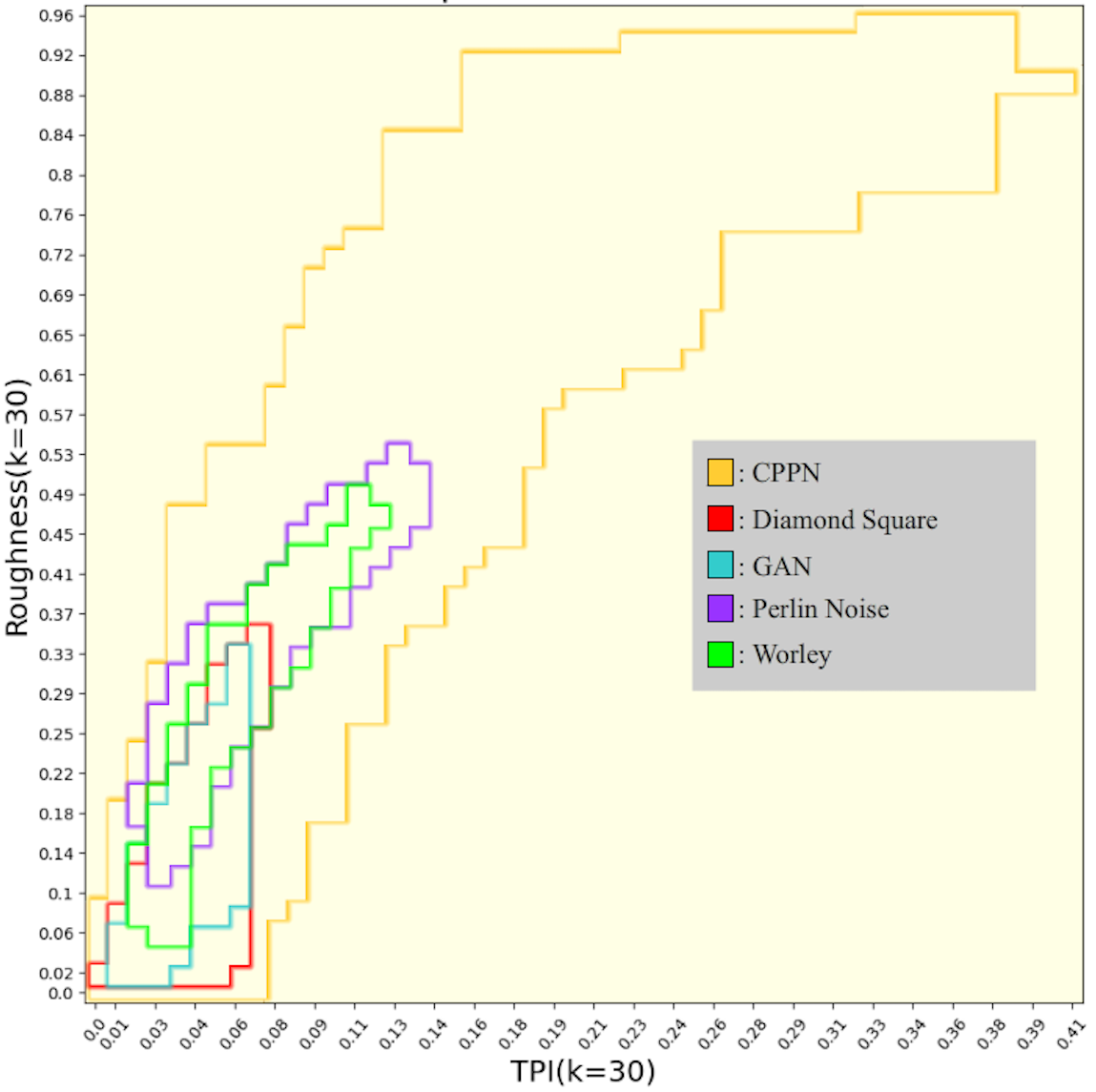}
\caption[]{Feature coverage of each generator after each experiment.}
\label{fig:fig5}
\end{figure}

\begin{figure*}[ht!]
\centering 
\includegraphics[width=2\columnwidth]{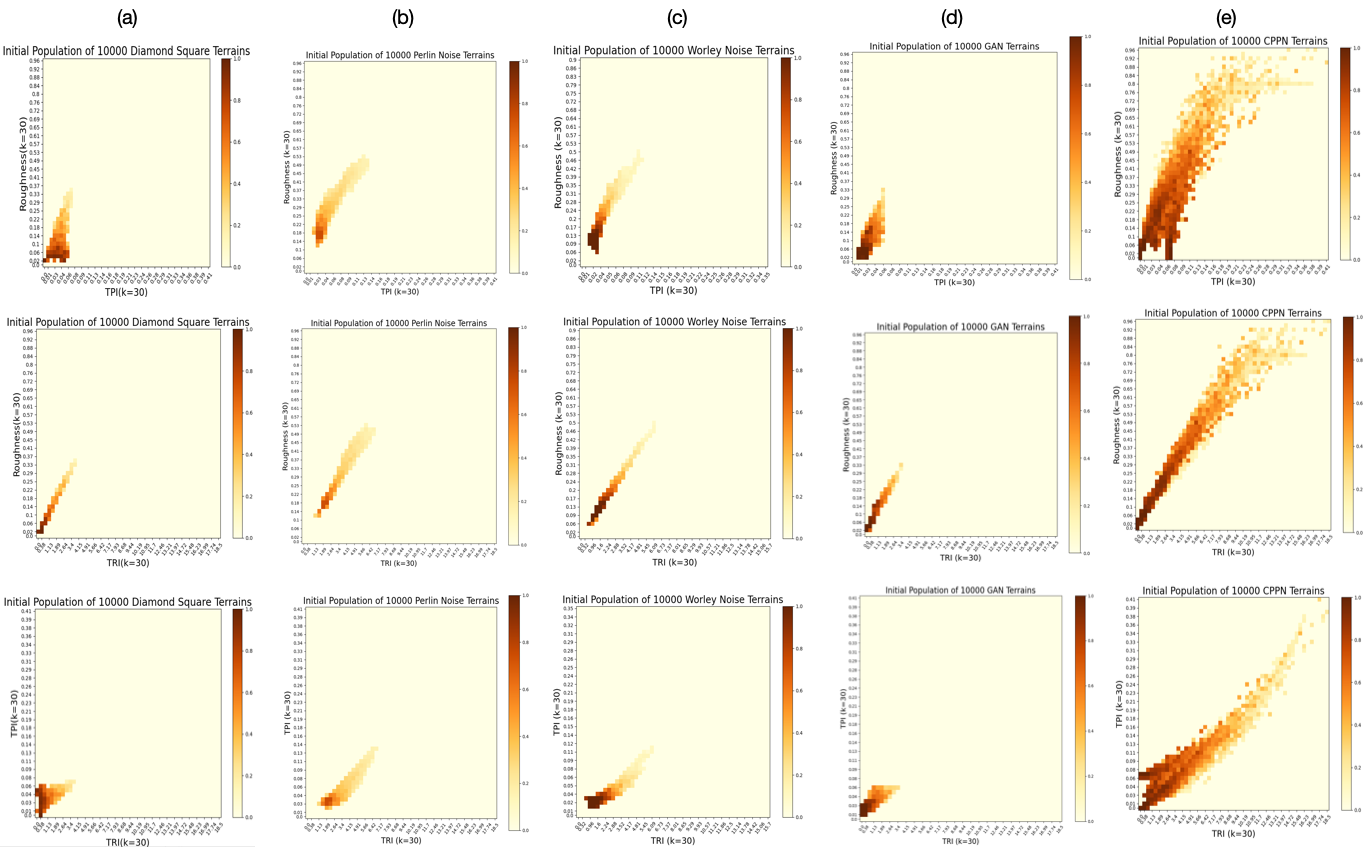}
\caption[]{MAP-Elites archive coverage for three combinations of features: TPI/TRI (row 1), roughness/TRI (row 2), roughness/TPI (row 3) for Diamond Square noise (a), Perlin noise (b), Worley noise (c), GAN (d) and CPPN (e) terrain generators after evolution. Colour corresponds to fitness, measured as the average performance of the base policy humanoid traversing each terrain.  Dark colours represent fitter behaviours.}
\label{fig:fig6}
\end{figure*}

The coverage of each terrain generator is shown in Figure \ref{fig:fig5}.  Pairwise feature comparisons are shown in Figure \ref{fig:fig6}.  All coverage maps show a strong gradient in terrain traversability, with it declining as the map feature values increase. This suggests the chosen features are very representative of traversability. Some variation exists between the different features, but the most significant basis of variation is in the euclidean distance between feature values and (0,0). Each terrain generator presents a characteristic shape and amount of coverage on the map. Table 1 shows the comparative coverage between generators, with CPPN having significantly higher coverage, and able to create more difficult and more diverse terrains than the other generators.    Interestingly the other indirect emitter, GAN, presents the lowest coverage of all. The GAN was trained on real terrains and therefore can only output similar meshes.  The CPPN is unrestricted in this manner.  Perlin noise shows the strongest diagonal mapping to the feature dimensions and presents the best direct encoding, followed by Worley noise.  Diamond-square noise can be seen to struggle to generate reasonable values for TPI.  Feature ranges were identical between generator coverage maps for fair comparison. While the total terrains in each map vary significantly based on coverage, adjusting feature ranges to account for this does not help learning as terrains with comparable features are similarly traversable and are skipped. 

\begin{table}[h!]
\centering
\caption{Coverage percent of total map for feature combinations with each terrain generator; maps from Figure \ref{fig:fig5}.}
\begin{tabular}{|p{2.3cm}| c c c|}
      \hline
 Terrain Generator & Roughness/TPI & Roughness/TRI & TPI/TRI \\ [0.5ex] 
 \hline
 CPPN & 23.16\% & 15.52\% & 14.92\% \\ 
 \hline
 GAN & 2.48\% & 2.48\% & 1.48\% \\
 \hline
 Perlin Noise & 3.92\% & 3.92\% & 2.80\% \\
 \hline
 Worley Noise & 2.88\% & 1.84\% & 2.12\% \\
 \hline
 Diamond Square & 3.12\% & 1.36\% & 1.88\% \\
 \hline \hline
 Combined & 23.64\% & 15.56\% & 15.56\% \\ [1ex] 
 \hline
\end{tabular}

\label{table:1}
\end{table}

\subsection{Experiment 2: Bipedal Walker Learning}

\begin{figure}[ht!]
\centering
\subfloat[]{\includegraphics[width=1.0\columnwidth]{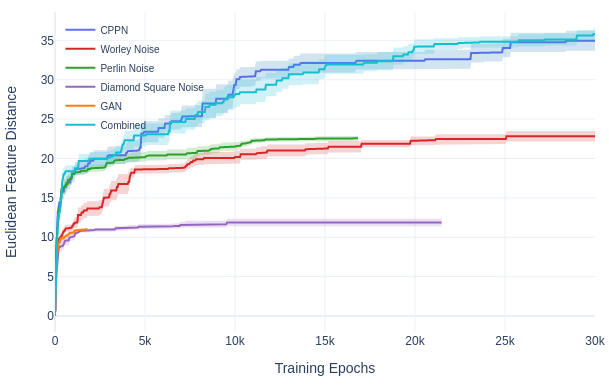}}\\
\subfloat[]{\includegraphics[width=1.0\columnwidth]{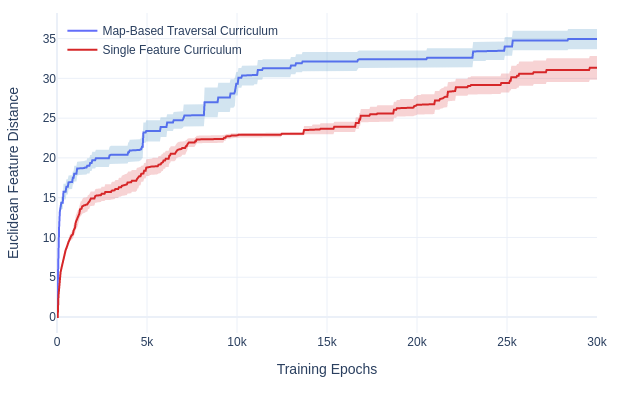}}
\caption[]{Illustrative performance comparison between (a) the different terrain generators, and (b) CPPN terrains between the proposed map-based curriculum algorithm and incrementally increasing roughness with standard CL, showing the most difficult terrain completed at the current training iteration. Plots finish when 30k epochs is reached or IT\&E has excluded all terrains. Averaged over 10 trials, with confidence shown for $\pm$1 standard error.}
\label{fig:fig7}
\end{figure}

In the second experiment we carry out a typical learning experiment on the evolved curricula.  We trained the walker from the base policy using the PyTorch PPO implementation from SpinningUp \cite{SpinningUp2018} with default parameters with a maximum of 30,000 epochs.  The base policy provides the prior for IT\&E.  Traversal of the map is the same Gaussian Process (GP) technique as described by Cully et al. \cite{gaussianprocess}, using the same recommended parameters: $\rho$ = 0.4, $\alpha$ = 0.9, $\kappa$ = 0.05, ${\sigma^2}_{noise}$ = 0.001, and the same covariance function (Matern kernel). PPO trains for 40 epochs at a time until the humanoid reaches the 90\% traversability threshold (90\% of maximum fitness achieved) or when there had been 100 traversability evaluations on the terrain. Once this condition is met, the GP model finds the highest fitness (easiest) terrain from evaluations with the updated PPO model. However, unlike the original IT\&E, our algorithm removes terrains that are estimated by the model to be easier than the terrain just trained on at each update. This ensures that the algorithm incrementally increases difficulty. This also avoids the problem of the curriculum visiting previously easy terrains that may appear hard after learning on complex environments which require a significant change in policy. The learning process ends when all terrains are used. To assess the performance of each terrain generator's curriculum map, the hardest terrain successfully traversed at each training epoch was recorded. This is shown in Figure 7, and includes training using a combined map with the highest fitness terrain chosen when grid squares overlap between generators. Hardest terrains are measured in terms of how large their feature values are, which is the euclidean feature distance from (0,0). CPPN and Combined significantly outperform the other four generators with very similar large coverage profiles. CPPN and combined are similar in performance.  Diamond Square Noise and the GAN perform similarly, and have similar coverage maps. Worley and Perlin Noise also obtain a similar final difficulty, and have similarly shaped coverage maps although Perlin covers more of the archive. Learning speed is another key determinant of generator performance --- Fig.\ref{fig:fig7}. We measure this by designating terrains at a certain Euclidean distance from the top-right corner of the map, and recording the iteration at which the walker learns one of those terrains.  Table \ref{table:2} shows some interesting results.  First, the Perlin curriculum learns much faster than the Worley curriculum, despite covering comparable archive area.  The high regularity of Perlin Noise may account for this with policies that are more generalisable between Perlin Noise terrains.  Second, although CPPN and combined maps achieve comparable difficulty, a combined map that starts on higher-fitness noise terrains before transitioning to CPPN terrains achieves difficulty milestones much faster than the single CPPN generator.  The result suggests that combining different generators in a single curriculum may be beneficial, however investigating is out of scope for the current study.  Finally, the effect of the map creation and map-traversal algorithm on performance against classic curriculum learning with a single feature (Figure \ref{fig:fig7}(b)). CPPN terrains from the map were sorted by their roughness, and every fifth terrain was used, similar to standard CL by incrementally increasing a difficulty parameter. This trained much slower and did not succeed on terrains as difficult as in the map-based algorithm.

\begin{table*}[ht!]
\centering
\caption{Showing learning speed: the number of training epochs required to learn a designated terrain a given \% away from the maximum possible map distance (70.71).}
\begin{tabular}{|p{2.3cm}| c c c c c c c c c c|}
      \hline
 Terrain Generator & 5\% & 10\% & 15\% & 20\% & 25\% & 30\% & 35\% & 40\% & 45\% & 50\% \\ [5.0ex] 
 \hline
 CPPN & \textbf{40} & 80 & \textbf{120} & \textbf{240} & 960 & 4720 & \textbf{7000} &  \textbf{9880} & \textbf{13280} & -  \\ 
 \hline
 GAN & \textbf{40} & 80 & 1000 & - & - & - & - & - & - & - \\
 \hline 
 Perlin Noise & \textbf{40} & \textbf{40} & 160 & 320 & 960 & 8680 & - & - & - & -  \\
 \hline
 Worley Noise & \textbf{40} & 80 & 480 & 2640 & 4160 & 14080 & - & - & - & - \\
 \hline
 Diamond Square & 80 & 160 & 1280 & - & - & - & - & - & - & - \\
 \hline \hline
 Combined & 80 & 120 & 160 & 320 & \textbf{440} & \textbf{3720} & 7280 & 10280 & 15040 & \textbf{29000} \\ [1ex] 
 \hline

\end{tabular}
\label{table:2}
\end{table*}



\section{Discussion}

In this paper we investigated the utility of different feature descriptors for map-based curricula.  Using a ground truth of how hard a terrain was for a pre-trained bipedal walker, we showed how several feature descriptors taken from the literature allow for the assessment of meaningful difficulty purely from a terrain mesh.  Using pairwise combinations of features as MAP-Elites feature dimensions, we then showed the effect of feature dimension selection on archive coverage and fill pattern. 

In a second step, we used our selected feature descriptors to build a MAP-Elites-based curriculum for different terrain generation algorithms, including direct and indirect representation.  Using our representation-agnostic feature descriptors, we could directly compare archive properties.  We show that CPPNs present the best coverage, and make the most difficult terrains learnable by the agent. GANs were most limited in terms of coverage, with noise generators taking up the middle ground.  Perlin and Worley noise in particular presented useful coverage patterns. Interestingly, Perlin noise allowed for much faster learning (achievement of terrain milestones) than Worley.  Even more interestingly, we see that a combined map presents significantly faster learning than a pure CPPN map, as it learns first on fitter (easier) terrains (where cells are shared with multiple generators) before transitioning to pure CPPN terrains that the other generators cannot reach.

Future work will investigate this, as well as exploring the effects of hyperparameter tuning (all parameters were set to default), as well as the generalisability of these findings to other high-impact problems.  However, purely in terms of learning about how agent-based curriculum RL works, this paper presents several important contributions for the research community.  Future work will also apply the methods presented to the training of real robots, where the effect of generator types and the map-based curriculum may be able to contribute in narrowing the sim2real gap in addition to increasing learning speed. Moreover, game-playing agents could benefit from this curriculum technique. Further exploration of feature descriptors should be conducted in different curriculum contexts to measure their influence on policy performance. We hope this will lead to more terrain generator methods used in curriculum learning, as well as more principled selection of generator.

\bibliographystyle{ACM-Reference-Format}
\bibliography{main}


\begin{thebibliography}{38}


\ifx \showCODEN    \undefined \def \showCODEN     #1{\unskip}     \fi
\ifx \showDOI      \undefined \def \showDOI       #1{#1}\fi
\ifx \showISBNx    \undefined \def \showISBNx     #1{\unskip}     \fi
\ifx \showISBNxiii \undefined \def \showISBNxiii  #1{\unskip}     \fi
\ifx \showISSN     \undefined \def \showISSN      #1{\unskip}     \fi
\ifx \showLCCN     \undefined \def \showLCCN      #1{\unskip}     \fi
\ifx \shownote     \undefined \def \shownote      #1{#1}          \fi
\ifx \showarticletitle \undefined \def \showarticletitle #1{#1}   \fi
\ifx \showURL      \undefined \def \showURL       {\relax}        \fi
\providecommand\bibfield[2]{#2}
\providecommand\bibinfo[2]{#2}
\providecommand\natexlab[1]{#1}
\providecommand\showeprint[2][]{arXiv:#2}

\bibitem[\protect\citeauthoryear{A.~Cully and J.Mouret}{A.~Cully and
  J.Mouret}{2015}]%
        {gaussianprocess}
\bibfield{author}{\bibinfo{person}{D.~Tarapore A.~Cully, J.~Clune} {and}
  \bibinfo{person}{J.Mouret}.} \bibinfo{year}{2015}\natexlab{}.
\newblock \showarticletitle{Robots that can adapt like animals}.
\newblock \bibinfo{journal}{\emph{Nature}} (\bibinfo{year}{2015}).
\newblock
\urldef\tempurl%
\url{https://doi.org/10.1038/nature14422}
\showDOI{\tempurl}


\bibitem[\protect\citeauthoryear{Achiam}{Achiam}{2018}]%
        {SpinningUp2018}
\bibfield{author}{\bibinfo{person}{Joshua Achiam}.}
  \bibinfo{year}{2018}\natexlab{}.
\newblock \showarticletitle{{Spinning Up in Deep Reinforcement Learning}}.
\newblock  (\bibinfo{year}{2018}).
\newblock


\bibitem[\protect\citeauthoryear{Akkaya, Andrychowicz, Chociej, Litwin, McGrew,
  Petron, Paino, Plappert, Powell, Ribas, et~al\mbox{.}}{Akkaya
  et~al\mbox{.}}{2019}]%
        {akkaya2019solving}
\bibfield{author}{\bibinfo{person}{Ilge Akkaya}, \bibinfo{person}{Marcin
  Andrychowicz}, \bibinfo{person}{Maciek Chociej}, \bibinfo{person}{Mateusz
  Litwin}, \bibinfo{person}{Bob McGrew}, \bibinfo{person}{Arthur Petron},
  \bibinfo{person}{Alex Paino}, \bibinfo{person}{Matthias Plappert},
  \bibinfo{person}{Glenn Powell}, \bibinfo{person}{Raphael Ribas},
  {et~al\mbox{.}}} \bibinfo{year}{2019}\natexlab{}.
\newblock \showarticletitle{Solving rubik's cube with a robot hand}.
\newblock \bibinfo{journal}{\emph{arXiv preprint arXiv:1910.07113}}
  (\bibinfo{year}{2019}).
\newblock


\bibitem[\protect\citeauthoryear{Alvarez, Font~Fernandez, Dahlskog, and
  Togelius}{Alvarez et~al\mbox{.}}{2020}]%
        {9300206}
\bibfield{author}{\bibinfo{person}{Alberto Alvarez}, \bibinfo{person}{Jose
  Maria~Maria Font~Fernandez}, \bibinfo{person}{Steve Dahlskog}, {and}
  \bibinfo{person}{Julian Togelius}.} \bibinfo{year}{2020}\natexlab{}.
\newblock \showarticletitle{Interactive Constrained MAP-Elites: Analysis and
  Evaluation of the Expressiveness of the Feature Dimensions}.
\newblock \bibinfo{journal}{\emph{IEEE Transactions on Games}}
  (\bibinfo{year}{2020}), \bibinfo{pages}{1--1}.
\newblock
\urldef\tempurl%
\url{https://doi.org/10.1109/TG.2020.3046133}
\showDOI{\tempurl}


\bibitem[\protect\citeauthoryear{Bengio, Louradour, Collobert, and
  Weston}{Bengio et~al\mbox{.}}{2009}]%
        {bengio2009curriculum}
\bibfield{author}{\bibinfo{person}{Yoshua Bengio},
  \bibinfo{person}{J{\'e}r{\^o}me Louradour}, \bibinfo{person}{Ronan
  Collobert}, {and} \bibinfo{person}{Jason Weston}.}
  \bibinfo{year}{2009}\natexlab{}.
\newblock \showarticletitle{Curriculum learning}. In
  \bibinfo{booktitle}{\emph{Proceedings of the 26th annual international
  conference on machine learning}}. \bibinfo{pages}{41--48}.
\newblock


\bibitem[\protect\citeauthoryear{Bongard}{Bongard}{[n.\,d.]}]%
        {bongard_utility_2010}
\bibfield{author}{\bibinfo{person}{Josh Bongard}.}
  \bibinfo{year}{[n.\,d.]}\natexlab{}.
\newblock \showarticletitle{The Utility of Evolving Simulated Robot Morphology
  Increases with Task Complexity for Object Manipulation}.
\newblock  \bibinfo{volume}{16}, \bibinfo{number}{3}
  (\bibinfo{year}{[n.\,d.]}), \bibinfo{pages}{201--223}.
\newblock
\showISSN{1064-5462}
\urldef\tempurl%
\url{https://doi.org/10.1162/artl.2010.Bongard.024}
\showDOI{\tempurl}
\newblock
\shownote{Conference Name: Artificial Life}.


\bibitem[\protect\citeauthoryear{Chavez-Garcia, Guzzi, Gambardella, and
  Giusti}{Chavez-Garcia et~al\mbox{.}}{2018}]%
        {traversability}
\bibfield{author}{\bibinfo{person}{R.~Omar Chavez-Garcia},
  \bibinfo{person}{Jérôme Guzzi}, \bibinfo{person}{Luca~M. Gambardella},
  {and} \bibinfo{person}{Alessandro Giusti}.} \bibinfo{year}{2018}\natexlab{}.
\newblock \showarticletitle{Learning Ground Traversability From Simulations}.
\newblock \bibinfo{journal}{\emph{IEEE Robotics and Automation Letters}}
  \bibinfo{volume}{3}, \bibinfo{number}{3} (\bibinfo{year}{2018}),
  \bibinfo{pages}{1695--1702}.
\newblock
\urldef\tempurl%
\url{https://doi.org/10.1109/LRA.2018.2801794}
\showDOI{\tempurl}


\bibitem[\protect\citeauthoryear{Coumans and Bai}{Coumans and Bai}{2017}]%
        {coumans2017pybullet}
\bibfield{author}{\bibinfo{person}{Erwin Coumans} {and} \bibinfo{person}{Yunfei
  Bai}.} \bibinfo{year}{2017}\natexlab{}.
\newblock \bibinfo{title}{Pybullet, a python module for physics simulation in
  robotics, games and machine learning}.
\newblock
\newblock


\bibitem[\protect\citeauthoryear{Cully, Clune, Tarapore, and Mouret}{Cully
  et~al\mbox{.}}{2015}]%
        {Cully_2015}
\bibfield{author}{\bibinfo{person}{Antoine Cully}, \bibinfo{person}{Jeff
  Clune}, \bibinfo{person}{Danesh Tarapore}, {and}
  \bibinfo{person}{Jean-Baptiste Mouret}.} \bibinfo{year}{2015}\natexlab{}.
\newblock \showarticletitle{Robots that can adapt like animals}.
\newblock \bibinfo{journal}{\emph{Nature}} \bibinfo{volume}{521},
  \bibinfo{number}{7553} (\bibinfo{date}{May} \bibinfo{year}{2015}),
  \bibinfo{pages}{503–507}.
\newblock
\showISSN{1476-4687}
\urldef\tempurl%
\url{https://doi.org/10.1038/nature14422}
\showDOI{\tempurl}


\bibitem[\protect\citeauthoryear{{De Reu}, Bourgeois, Bats, Zwertvaegher,
  Gelorini, {De Smedt}, Chu, Antrop, {De Maeyer}, Finke, {Van Meirvenne},
  Verniers, and Crombé}{{De Reu} et~al\mbox{.}}{2013}]%
        {tpi}
\bibfield{author}{\bibinfo{person}{Jeroen {De Reu}}, \bibinfo{person}{Jean
  Bourgeois}, \bibinfo{person}{Machteld Bats}, \bibinfo{person}{Ann
  Zwertvaegher}, \bibinfo{person}{Vanessa Gelorini}, \bibinfo{person}{Philippe
  {De Smedt}}, \bibinfo{person}{Wei Chu}, \bibinfo{person}{Marc Antrop},
  \bibinfo{person}{Philippe {De Maeyer}}, \bibinfo{person}{Peter Finke},
  \bibinfo{person}{Marc {Van Meirvenne}}, \bibinfo{person}{Jacques Verniers},
  {and} \bibinfo{person}{Philippe Crombé}.} \bibinfo{year}{2013}\natexlab{}.
\newblock \showarticletitle{Application of the topographic position index to
  heterogeneous landscapes}.
\newblock \bibinfo{journal}{\emph{Geomorphology}}  \bibinfo{volume}{186}
  (\bibinfo{year}{2013}), \bibinfo{pages}{39--49}.
\newblock
\showISSN{0169-555X}
\urldef\tempurl%
\url{https://doi.org/10.1016/j.geomorph.2012.12.015}
\showDOI{\tempurl}


\bibitem[\protect\citeauthoryear{Deng, Wilson, and Bauer}{Deng
  et~al\mbox{.}}{2007}]%
        {roughness}
\bibfield{author}{\bibinfo{person}{Y. Deng}, \bibinfo{person}{J.~P. Wilson},
  {and} \bibinfo{person}{B.~O. Bauer}.} \bibinfo{year}{2007}\natexlab{}.
\newblock \showarticletitle{DEM resolution dependencies of terrain attributes
  across a landscape}.
\newblock \bibinfo{journal}{\emph{International Journal of Geographical
  Information Science}} \bibinfo{volume}{21}, \bibinfo{number}{2}
  (\bibinfo{year}{2007}), \bibinfo{pages}{187--213}.
\newblock
\urldef\tempurl%
\url{https://doi.org/10.1080/13658810600894364}
\showDOI{\tempurl}
\showeprint{https://doi.org/10.1080/13658810600894364}


\bibitem[\protect\citeauthoryear{Florensa, Held, Geng, and Abbeel}{Florensa
  et~al\mbox{.}}{2018}]%
        {florensa2018automatic}
\bibfield{author}{\bibinfo{person}{Carlos Florensa}, \bibinfo{person}{David
  Held}, \bibinfo{person}{Xinyang Geng}, {and} \bibinfo{person}{Pieter
  Abbeel}.} \bibinfo{year}{2018}\natexlab{}.
\newblock \showarticletitle{Automatic goal generation for reinforcement
  learning agents}. In \bibinfo{booktitle}{\emph{International conference on
  machine learning}}. PMLR, \bibinfo{pages}{1515--1528}.
\newblock


\bibitem[\protect\citeauthoryear{Fournier, Fussell, and Carpenter}{Fournier
  et~al\mbox{.}}{1982}]%
        {diamondsq}
\bibfield{author}{\bibinfo{person}{Alain Fournier}, \bibinfo{person}{Don
  Fussell}, {and} \bibinfo{person}{Loren Carpenter}.}
  \bibinfo{year}{1982}\natexlab{}.
\newblock \showarticletitle{Computer Rendering of Stochastic Models}.
\newblock \bibinfo{journal}{\emph{Commun. ACM}} \bibinfo{volume}{25},
  \bibinfo{number}{6} (\bibinfo{date}{June} \bibinfo{year}{1982}),
  \bibinfo{pages}{371–384}.
\newblock
\showISSN{0001-0782}
\urldef\tempurl%
\url{https://doi.org/10.1145/358523.358553}
\showDOI{\tempurl}


\bibitem[\protect\citeauthoryear{Frade, Vega, and Cotta}{Frade
  et~al\mbox{.}}{2009}]%
        {evoTG}
\bibfield{author}{\bibinfo{person}{Miguel Frade}, \bibinfo{person}{Francisco
  Vega}, {and} \bibinfo{person}{Carlos Cotta}.}
  \bibinfo{year}{2009}\natexlab{}.
\newblock \showarticletitle{Breeding Terrains with Genetic Terrain Programming:
  The Evolution of Terrain Generators}.
\newblock \bibinfo{journal}{\emph{Int. J. Computer Games Technology}}
  \bibinfo{volume}{2009} (\bibinfo{date}{12} \bibinfo{year}{2009}).
\newblock
\urldef\tempurl%
\url{https://doi.org/10.1155/2009/125714}
\showDOI{\tempurl}


\bibitem[\protect\citeauthoryear{Heess, TB, Sriram, Lemmon, Merel, Wayne,
  Tassa, Erez, Wang, Eslami, Riedmiller, and Silver}{Heess
  et~al\mbox{.}}{2017}]%
        {heess2017emergence}
\bibfield{author}{\bibinfo{person}{Nicolas Heess}, \bibinfo{person}{Dhruva TB},
  \bibinfo{person}{Srinivasan Sriram}, \bibinfo{person}{Jay Lemmon},
  \bibinfo{person}{Josh Merel}, \bibinfo{person}{Greg Wayne},
  \bibinfo{person}{Yuval Tassa}, \bibinfo{person}{Tom Erez},
  \bibinfo{person}{Ziyu Wang}, \bibinfo{person}{S.~M.~Ali Eslami},
  \bibinfo{person}{Martin Riedmiller}, {and} \bibinfo{person}{David Silver}.}
  \bibinfo{year}{2017}\natexlab{}.
\newblock \bibinfo{title}{Emergence of Locomotion Behaviours in Rich
  Environments}.
\newblock
\newblock
\showeprint[arxiv]{1707.02286}~[cs.AI]


\bibitem[\protect\citeauthoryear{Hofer, Bekris, Handa, Gamboa, Mozifian,
  Golemo, Atkeson, Fox, Goldberg, Leonard, Karen~Liu, Peters, Song, Welinder,
  and White}{Hofer et~al\mbox{.}}{2021}]%
        {HoferSebastian2021SiRa}
\bibfield{author}{\bibinfo{person}{Sebastian Hofer}, \bibinfo{person}{Kostas
  Bekris}, \bibinfo{person}{Ankur Handa}, \bibinfo{person}{Juan~Camilo Gamboa},
  \bibinfo{person}{Melissa Mozifian}, \bibinfo{person}{Florian Golemo},
  \bibinfo{person}{Chris Atkeson}, \bibinfo{person}{Dieter Fox},
  \bibinfo{person}{Ken Goldberg}, \bibinfo{person}{John Leonard},
  \bibinfo{person}{C Karen~Liu}, \bibinfo{person}{Jan Peters},
  \bibinfo{person}{Shuran Song}, \bibinfo{person}{Peter Welinder}, {and}
  \bibinfo{person}{Martha White}.} \bibinfo{year}{2021}\natexlab{}.
\newblock \showarticletitle{Sim2Real in Robotics and Automation: Applications
  and Challenges}.
\newblock \bibinfo{journal}{\emph{IEEE transactions on automation science and
  engineering}} \bibinfo{volume}{18}, \bibinfo{number}{2}
  (\bibinfo{year}{2021}), \bibinfo{pages}{398--400}.
\newblock
\showISSN{1545-5955}


\bibitem[\protect\citeauthoryear{Howard and Elfes}{Howard and Elfes}{2014}]%
        {howard2014evolving}
\bibfield{author}{\bibinfo{person}{David Howard} {and} \bibinfo{person}{Alberto
  Elfes}.} \bibinfo{year}{2014}\natexlab{}.
\newblock \showarticletitle{Evolving spiking networks for turbulence-tolerant
  quadrotor control}. In \bibinfo{booktitle}{\emph{ALIFE 14: The Fourteenth
  International Conference on the Synthesis and Simulation of Living Systems}}.
  MIT Press, \bibinfo{pages}{431--438}.
\newblock


\bibitem[\protect\citeauthoryear{Huizinga and Clune}{Huizinga and
  Clune}{2019}]%
        {huizinga2019evolving}
\bibfield{author}{\bibinfo{person}{Joost Huizinga} {and} \bibinfo{person}{Jeff
  Clune}.} \bibinfo{year}{2019}\natexlab{}.
\newblock \bibinfo{title}{Evolving Multimodal Robot Behavior via Many Stepping
  Stones with the Combinatorial Multi-Objective Evolutionary Algorithm}.
\newblock
\newblock
\showeprint[arxiv]{1807.03392}~[cs.NE]


\bibitem[\protect\citeauthoryear{Kadian, Truong, Gokaslan, Clegg, Wijmans, Lee,
  Savva, Chernova, and Batra}{Kadian et~al\mbox{.}}{2020}]%
        {KadianAbhishek2020SPDE}
\bibfield{author}{\bibinfo{person}{Abhishek Kadian}, \bibinfo{person}{Joanne
  Truong}, \bibinfo{person}{Aaron Gokaslan}, \bibinfo{person}{Alexander Clegg},
  \bibinfo{person}{Erik Wijmans}, \bibinfo{person}{Stefan Lee},
  \bibinfo{person}{Manolis Savva}, \bibinfo{person}{Sonia Chernova}, {and}
  \bibinfo{person}{Dhruv Batra}.} \bibinfo{year}{2020}\natexlab{}.
\newblock \showarticletitle{Sim2Real Predictivity: Does Evaluation in
  Simulation Predict Real-World Performance?}
\newblock \bibinfo{journal}{\emph{IEEE robotics and automation letters}}
  \bibinfo{volume}{5}, \bibinfo{number}{4} (\bibinfo{year}{2020}),
  \bibinfo{pages}{6670--6677}.
\newblock
\showISSN{2377-3766}


\bibitem[\protect\citeauthoryear{Kober, Bagnell, and Peters}{Kober
  et~al\mbox{.}}{2013}]%
        {kober2013reinforcement}
\bibfield{author}{\bibinfo{person}{Jens Kober}, \bibinfo{person}{J~Andrew
  Bagnell}, {and} \bibinfo{person}{Jan Peters}.}
  \bibinfo{year}{2013}\natexlab{}.
\newblock \showarticletitle{Reinforcement learning in robotics: A survey}.
\newblock \bibinfo{journal}{\emph{The International Journal of Robotics
  Research}} \bibinfo{volume}{32}, \bibinfo{number}{11} (\bibinfo{year}{2013}),
  \bibinfo{pages}{1238--1274}.
\newblock


\bibitem[\protect\citeauthoryear{Lee, Hwangbo, Wellhausen, Koltun, and
  Hutter}{Lee et~al\mbox{.}}{2020}]%
        {lee_learning_2020}
\bibfield{author}{\bibinfo{person}{Joonho Lee}, \bibinfo{person}{Jemin
  Hwangbo}, \bibinfo{person}{Lorenz Wellhausen}, \bibinfo{person}{Vladlen
  Koltun}, {and} \bibinfo{person}{Marco Hutter}.}
  \bibinfo{year}{2020}\natexlab{}.
\newblock \showarticletitle{Learning quadrupedal locomotion over challenging
  terrain}.
\newblock \bibinfo{journal}{\emph{Science Robotics}} \bibinfo{volume}{5},
  \bibinfo{number}{47} (\bibinfo{year}{2020}).
\newblock
\urldef\tempurl%
\url{https://doi.org/10.1126/scirobotics.abc5986}
\showDOI{\tempurl}


\bibitem[\protect\citeauthoryear{Miras and Eiben}{Miras and Eiben}{[n.\,d.]}]%
        {miras_effects_2019}
\bibfield{author}{\bibinfo{person}{Karine Miras} {and} \bibinfo{person}{A.~E.
  Eiben}.} \bibinfo{year}{[n.\,d.]}\natexlab{}.
\newblock \showarticletitle{Effects of environmental conditions on evolved
  robot morphologies and behavior}. In \bibinfo{booktitle}{\emph{Proceedings of
  the Genetic and Evolutionary Computation Conference}} (Prague Czech Republic,
  2019-07-13). \bibinfo{publisher}{{ACM}}, \bibinfo{pages}{125--132}.
\newblock
\showISBNx{978-1-4503-6111-8}
\urldef\tempurl%
\url{https://doi.org/10.1145/3321707.3321811}
\showDOI{\tempurl}


\bibitem[\protect\citeauthoryear{Mouret and Clune}{Mouret and Clune}{2015}]%
        {mouret2015illuminating}
\bibfield{author}{\bibinfo{person}{Jean-Baptiste Mouret} {and}
  \bibinfo{person}{Jeff Clune}.} \bibinfo{year}{2015}\natexlab{}.
\newblock \bibinfo{title}{Illuminating search spaces by mapping elites}.
\newblock
\newblock
\showeprint[arxiv]{1504.04909}~[cs.AI]


\bibitem[\protect\citeauthoryear{Narvekar, Peng, Leonetti, Sinapov, Taylor, and
  Stone}{Narvekar et~al\mbox{.}}{2020}]%
        {narvekar2020curriculum}
\bibfield{author}{\bibinfo{person}{Sanmit Narvekar}, \bibinfo{person}{Bei
  Peng}, \bibinfo{person}{Matteo Leonetti}, \bibinfo{person}{Jivko Sinapov},
  \bibinfo{person}{Matthew~E Taylor}, {and} \bibinfo{person}{Peter Stone}.}
  \bibinfo{year}{2020}\natexlab{}.
\newblock \showarticletitle{Curriculum learning for reinforcement learning
  domains: A framework and survey}.
\newblock \bibinfo{journal}{\emph{arXiv preprint arXiv:2003.04960}}
  (\bibinfo{year}{2020}).
\newblock


\bibitem[\protect\citeauthoryear{Nichol, Pfau, Hesse, Klimov, and
  Schulman}{Nichol et~al\mbox{.}}{2018}]%
        {nichol2018gotta}
\bibfield{author}{\bibinfo{person}{Alex Nichol}, \bibinfo{person}{Vicki Pfau},
  \bibinfo{person}{Christopher Hesse}, \bibinfo{person}{Oleg Klimov}, {and}
  \bibinfo{person}{John Schulman}.} \bibinfo{year}{2018}\natexlab{}.
\newblock \bibinfo{title}{Gotta Learn Fast: A New Benchmark for Generalization
  in RL}.
\newblock
\newblock
\showeprint[arxiv]{1804.03720}~[cs.LG]


\bibitem[\protect\citeauthoryear{Ong, Saunders, Keyser, and Leggett}{Ong
  et~al\mbox{.}}{2005}]%
        {asasas}
\bibfield{author}{\bibinfo{person}{TeongJoo Ong}, \bibinfo{person}{Ryan
  Saunders}, \bibinfo{person}{John Keyser}, {and} \bibinfo{person}{John
  Leggett}.} \bibinfo{year}{2005}\natexlab{}.
\newblock \showarticletitle{Terrain generation using genetic algorithms}.
  \bibinfo{pages}{1463--1470}.
\newblock
\urldef\tempurl%
\url{https://doi.org/10.1145/1068009.1068241}
\showDOI{\tempurl}


\bibitem[\protect\citeauthoryear{Pugh, Soros, and Stanley}{Pugh
  et~al\mbox{.}}{2016}]%
        {pugh2016quality}
\bibfield{author}{\bibinfo{person}{Justin~K Pugh}, \bibinfo{person}{Lisa~B
  Soros}, {and} \bibinfo{person}{Kenneth~O Stanley}.}
  \bibinfo{year}{2016}\natexlab{}.
\newblock \showarticletitle{Quality diversity: A new frontier for evolutionary
  computation}.
\newblock \bibinfo{journal}{\emph{Frontiers in Robotics and AI}}
  \bibinfo{volume}{3} (\bibinfo{year}{2016}), \bibinfo{pages}{40}.
\newblock


\bibitem[\protect\citeauthoryear{Qin, Gao, and Bai}{Qin et~al\mbox{.}}{2019}]%
        {9043822}
\bibfield{author}{\bibinfo{person}{Bangyu Qin}, \bibinfo{person}{Yue Gao},
  {and} \bibinfo{person}{Yi Bai}.} \bibinfo{year}{2019}\natexlab{}.
\newblock \showarticletitle{Sim-to-real: Six-legged Robot Control with Deep
  Reinforcement Learning and Curriculum Learning}. In
  \bibinfo{booktitle}{\emph{2019 4th International Conference on Robotics and
  Automation Engineering (ICRAE)}}. \bibinfo{pages}{1--5}.
\newblock
\urldef\tempurl%
\url{https://doi.org/10.1109/ICRAE48301.2019.9043822}
\showDOI{\tempurl}


\bibitem[\protect\citeauthoryear{Riley, Degloria, and Elliot}{Riley
  et~al\mbox{.}}{1999}]%
        {tri}
\bibfield{author}{\bibinfo{person}{Shawn Riley}, \bibinfo{person}{Stephen
  Degloria}, {and} \bibinfo{person}{S.D. Elliot}.}
  \bibinfo{year}{1999}\natexlab{}.
\newblock \showarticletitle{A Terrain Ruggedness Index that Quantifies
  Topographic Heterogeneity}.
\newblock \bibinfo{journal}{\emph{Internation Journal of Science}}
  \bibinfo{volume}{5} (\bibinfo{date}{01} \bibinfo{year}{1999}),
  \bibinfo{pages}{23--27}.
\newblock


\bibitem[\protect\citeauthoryear{Risi and Togelius}{Risi and Togelius}{2020}]%
        {risi2020increasing}
\bibfield{author}{\bibinfo{person}{Sebastian Risi} {and}
  \bibinfo{person}{Julian Togelius}.} \bibinfo{year}{2020}\natexlab{}.
\newblock \showarticletitle{Increasing generality in machine learning through
  procedural content generation}.
\newblock \bibinfo{journal}{\emph{Nature Machine Intelligence}}
  \bibinfo{volume}{2}, \bibinfo{number}{8} (\bibinfo{year}{2020}),
  \bibinfo{pages}{428--436}.
\newblock


\bibitem[\protect\citeauthoryear{Rudin, Hoeller, Reist, and Hutter}{Rudin
  et~al\mbox{.}}{2021}]%
        {rudin2021learning}
\bibfield{author}{\bibinfo{person}{Nikita Rudin}, \bibinfo{person}{David
  Hoeller}, \bibinfo{person}{Philipp Reist}, {and} \bibinfo{person}{Marco
  Hutter}.} \bibinfo{year}{2021}\natexlab{}.
\newblock \showarticletitle{Learning to Walk in Minutes Using Massively
  Parallel Deep Reinforcement Learning}.
\newblock \bibinfo{journal}{\emph{arXiv:2109.11978 [cs.RO]}}
  (\bibinfo{year}{2021}).
\newblock
\showeprint[arxiv]{2109.11978}~[cs.RO]


\bibitem[\protect\citeauthoryear{Schulman, Wolski, Dhariwal, Radford, and
  Klimov}{Schulman et~al\mbox{.}}{2017}]%
        {ppo}
\bibfield{author}{\bibinfo{person}{John Schulman}, \bibinfo{person}{Filip
  Wolski}, \bibinfo{person}{Prafulla Dhariwal}, \bibinfo{person}{Alec Radford},
  {and} \bibinfo{person}{Oleg Klimov}.} \bibinfo{year}{2017}\natexlab{}.
\newblock \showarticletitle{Proximal Policy Optimization Algorithms}.
\newblock  (\bibinfo{date}{07} \bibinfo{year}{2017}).
\newblock


\bibitem[\protect\citeauthoryear{Secretan, Beato, D'Ambrosio, Rodriguez,
  Campbell, Folsom-Kovarik, and Stanley}{Secretan et~al\mbox{.}}{2011}]%
        {picbreeder}
\bibfield{author}{\bibinfo{person}{Jimmy Secretan}, \bibinfo{person}{Nicholas
  Beato}, \bibinfo{person}{David~B D'Ambrosio}, \bibinfo{person}{Adelein
  Rodriguez}, \bibinfo{person}{Adam Campbell}, \bibinfo{person}{Jeremiah~T
  Folsom-Kovarik}, {and} \bibinfo{person}{Kenneth~O Stanley}.}
  \bibinfo{year}{2011}\natexlab{}.
\newblock \showarticletitle{Picbreeder: A Case Study in Collaborative
  Evolutionary Exploration of Design Space}.
\newblock \bibinfo{journal}{\emph{Evolutionary computation}}
  \bibinfo{volume}{19}, \bibinfo{number}{3} (\bibinfo{year}{2011}),
  \bibinfo{pages}{373--403}.
\newblock
\showISSN{1530-9304}


\bibitem[\protect\citeauthoryear{Tidd, Hudson, and Cosgun}{Tidd
  et~al\mbox{.}}{2020}]%
        {TiddBrendan2020GCLf}
\bibfield{author}{\bibinfo{person}{Brendan Tidd}, \bibinfo{person}{Nicolas
  Hudson}, {and} \bibinfo{person}{Akansel Cosgun}.}
  \bibinfo{year}{2020}\natexlab{}.
\newblock \showarticletitle{Guided Curriculum Learning for Walking Over Complex
  Terrain}.
\newblock  (\bibinfo{year}{2020}).
\newblock


\bibitem[\protect\citeauthoryear{Wang, Lehman, Clune, and Stanley}{Wang
  et~al\mbox{.}}{2019}]%
        {poet}
\bibfield{author}{\bibinfo{person}{Rui Wang}, \bibinfo{person}{Joel Lehman},
  \bibinfo{person}{Jeff Clune}, {and} \bibinfo{person}{Kenneth~O Stanley}.}
  \bibinfo{year}{2019}\natexlab{}.
\newblock \showarticletitle{Paired Open-Ended Trailblazer (POET): Endlessly
  Generating Increasingly Complex and Diverse Learning Environments and Their
  Solutions}.
\newblock  (\bibinfo{year}{2019}).
\newblock


\bibitem[\protect\citeauthoryear{Worley}{Worley}{1996}]%
        {worley}
\bibfield{author}{\bibinfo{person}{Steven Worley}.}
  \bibinfo{year}{1996}\natexlab{}.
\newblock \showarticletitle{A Cellular Texture Basis Function}. In
  \bibinfo{booktitle}{\emph{Proceedings of the 23rd Annual Conference on
  Computer Graphics and Interactive Techniques}}
  \emph{(\bibinfo{series}{SIGGRAPH '96})}. \bibinfo{publisher}{Association for
  Computing Machinery}, \bibinfo{address}{New York, NY, USA},
  \bibinfo{pages}{291–294}.
\newblock
\showISBNx{0897917464}
\urldef\tempurl%
\url{https://doi.org/10.1145/237170.237267}
\showDOI{\tempurl}


\bibitem[\protect\citeauthoryear{Xie, Ling, Kim, and van~de Panne}{Xie
  et~al\mbox{.}}{2020}]%
        {xie2020allsteps}
\bibfield{author}{\bibinfo{person}{Zhaoming Xie}, \bibinfo{person}{Hung~Yu
  Ling}, \bibinfo{person}{Nam~Hee Kim}, {and} \bibinfo{person}{Michiel van~de
  Panne}.} \bibinfo{year}{2020}\natexlab{}.
\newblock \bibinfo{title}{ALLSTEPS: Curriculum-driven Learning of Stepping
  Stone Skills}.
\newblock
\newblock
\showeprint[arxiv]{2005.04323}~[cs.GR]


\bibitem[\protect\citeauthoryear{Zhou and Vanschoren}{Zhou and
  Vanschoren}{2021}]%
        {zhou2021openended}
\bibfield{author}{\bibinfo{person}{Fangqin Zhou} {and} \bibinfo{person}{Joaquin
  Vanschoren}.} \bibinfo{year}{2021}\natexlab{}.
\newblock \showarticletitle{Open-Ended Learning Strategies for Learning Complex
  Locomotion Skills}. In \bibinfo{booktitle}{\emph{Fifth Workshop on
  Meta-Learning at the Conference on Neural Information Processing Systems}}.
\newblock
\urldef\tempurl%
\url{https://openreview.net/forum?id=l8c9NYgA4Lw}
\showURL{%
\tempurl}


\end{thebibliography}

\end{document}